\documentclass[11pt,twocolumn]{scrartcl}

\usepackage{casa_conf}	
\usepackage{graphicx}	
\usepackage{amsmath}
\usepackage{graphicx}
\usepackage{lscape}
\usepackage{booktabs}
\usepackage{caption}
\usepackage{epsfig}

\title{Content-aware Density Map for Crowd Counting and Density Estimation }

\author{M.~M.~Oghaz$^1$\\
         \and
         A.~Khadka$^1$\\
         \and
        V.~Argyriou$^1$\\
           \and
        P.~Remagnino$^1$\\
       \and
         $^1$Kingston University, London, United Kingdom\\
         \{m.maktabdaroghaz, a.khadka, vasileios.argyriou, p.remagnino\}@kingston.ac.uk \\
        }

\begin{document}

\maketitle

\begin{abstract}
Precise knowledge about the size of a crowd, its density and flow can provide valuable information for  safety and security applications, event planning, architectural design and to analyze consumer behavior. Creating a powerful machine learning model, to employ for such applications requires a large and highly accurate and reliable dataset. Unfortunately the existing crowd counting and density estimation benchmark datasets are not only limited in terms of their size, but also lack annotation, in general too time consuming to implement. This paper attempts to address this very issue through a content aware technique, uses combinations of Chan-Vese segmentation algorithm, two-dimensional Gaussian filter  and brute-force nearest neighbor search. The results shows that by simply replacing the commonly used density map generators with the proposed method, higher level of accuracy can be achieved using the existing state of the art models.

\end{abstract}



\section{Introduction}

The study of human behavior is a subject of great scientific interest and probably an inexhaustible source of research. One of the most cited and popular research topic in human behavior analysis is study of crowd features and characteristics. In recent years, crowd analysis has gained a lot of interest mainly due to it’s wide range of applications such as safety monitoring, disaster management, public spaces design, and intelligence gathering, especially in the congested scenes like arenas, shopping malls, and airports \cite{pelechano2005crowd, silverman2005crowd}. 

Crowd counting, localization and density estimation are crucial objectives of an automated crowd analysis system.  Accurate knowledge of the crowd size, location and density in a public space can provide valuable insight for tasks such as city planning, analyzing consumer shopping patterns as well as maintaining general crowd safety. Several studies attempt to produce an accurate estimation of the true number of people present in a crowded scene through density estimation. 

Deep learning has proven superior to classic computer vision and machine learning techniques that tend to struggle with the complexity of crowd counting and behavior analysis models.  \cite{marsden2016fully}.

Generally, crowd counting and density estimation approaches can be divided in two categories: detection-based methods (specific) and regression-based methods (holistic). Detection-based methods generally assume each person on the crowd can be detected and located individually based on its individual features and characteristics. These approaches are preferable in sparse crowd analysis where crowd occlusion is negligible. Holistic crowd counting and behavior analysis approaches utilize global crowd features and characteristics to estimate  crowd size, flow and density. These approaches are preferable in dense crowd analysis, where crowd occlusion is significant. Due to high amount of occlusions these approaches only utilize heads as deterministic feature \cite{ryan2015evaluation}.

However, crowd counting and density estimation is not a trivial task. Several key challenges such as  severe occlusions, poor illumination, camera perspective and highly dynamic environments further complicate crowd analysis. Moreover, poor quality of annotated data increases to complexity of crowd counting and behavior analysis in crowded environments. 
The existing crowd counting and density estimation benchmark datasets are not only limited in terms of the quantity, but also lack in terms of annotation strategy.  

In regression-based crowd counting and density estimation approaches, people heads are the only  visible body part in an image. Thus, these approaches use heads as the only discriminant feature. Meanwhile, the existing benchmark datasets such as UCF-CC-50 and ShanghaiTech  only provide people heads centroid pixel instead of masking the entire head region. Hence, the recreation of the ground truth head masks is accomplished through a static two-dimensional Gaussian filter or a dynamic two-dimensional Gaussian based on the $K$ nearest neighbors.    
However, the dynamic Gaussian approach based on proximity of the nearest neighbors mitigates the issue to some extent, but this technique is not content aware and incorporates significant amount of noise into ground truth data \cite{idrees2018composition, zhang2016single}.

In this regard, our study attempts to address the limitation of the existing crowd counting and density estimation benchmark datasets through a content aware annotation technique. It employs combinations of nearest neighbor algorithm and unsupervised segmentation to generate the ground truth head masks. The proposed technique first uses the brute-force nearest neighbor search to localize the nearest neighbor head point, then it identified the head boundaries using Chan-Vese segmentation algorithm and generates a two-dimensional Gaussian filter on that basis. 
We believe that by simply replacing the $kNN$/Gaussian based ground truth density maps in an existing state of the art network with the proposed content aware approach in this study, higher level of accuracy can be achieved.    

The rest of this paper is organized as following: section 2 summarizes the related work, section 3 describes the existing datasets and annotation strategies, section 4 presents the proposed methodology, section 5 presents the experimental results and finally section 6 concludes the findings of this research.

\section{Related Work}

Over the last decade, there have been several studies  to address the problem of crowd counting and density estimation using deep learning techniques. 

Liu \textit{et al.} \cite{liu2019denet} proposed a universal network for counting people in a crowd with varying density and scale. in this study the proposed network is composed of two components: a detection network (DNet) and an encoder-decoder estimation network (ENet). The input first run through DNet to detect and count individuals who can be segmented clearly. Then, ENet is utilized to estimate the density maps of the remaining areas, where the numbers of individuals cannot be detected. Modified version of Xception used as an encoder for feature extraction and a combination of dilated convolution and transposed convolution used as decoder. 
Authors attempted to address the variations in crowd density with two literally isolated deep networks which significantly slows down the process lacks  novelty. 

In another study, Mehta \textit{et al.} \cite{valloli2019w} proposed independent decoding reinforcement branch as a binary classifier which helps the network converge much earlier and also enables the network to estimate density maps with high Structural Similarity Index (SSIM). A joint loss strategy, the {\em binary cross entropy} (BCE) loss and {\em mean squared error} (MSE) loss used to train the network in an end to end fashion. They have used variation of the U-net model to generate the density maps. The proposed model shows notable improvements in recreation of the crowd density maps over the existing models. 

A study by Oh \textit{et al.} \cite{oh2019crowd} attempt to address the uncertainty estimation in the domain of crowd counting. This study proposed a scalable neural network framework with quantification of decomposed uncertainty using a bootstrap ensemble. The proposed method incorporates both epistemic uncertainty and aleatoric uncertainty in a neural network for crowd counting. The proposed uncertainty quantification method provides additional auxiliary insight to the crowd counting model. The proposed technique attempt to address the uncertainty issue in crowd counting. However the  use of unsupervised calibration method to re-calibrate the predictions of the pre-trained network is questionable. 

In another study Olmschenk \textit{et al.} \cite{olmschenk2019improving} attempt to address the inefficiency of the existing crowd density map labeling scheme for training deep neural networks. This study proposes a labeling scheme based on inverse k-nearest neighbor ($ikNN$) maps which does not explicitly represents the crowd density. Authors claim a single $ikNN$ map provides information similar to the commonly practiced accumulation of many density maps with different Gaussian spreads.

A study by Idrees \textit{et al.} \cite{idrees2018composition} stems from the observation that crowd counting, density map estimation and localization are very interrelated and can be decomposed with respect to each other through composition loss, which can then be used to train a neural network.
This study 

Several other studies including \cite{jiang2018mask, varior2019scale, liu2018leveraging, change2013semi, hossain2019crowd,  wang2019learning,  wang2018defense, kang2018crowd,  liu2018crowd, kim2018comparison, ranjan2018iterative, shi2018crowd, babu2018divide} attempted to address crowd counting, localization and density estimation issues yet majority of these approaches employed the flawed ground truth density map generation approach.

\section{Annotation Strategy}

In a dense crowd scenario, aside from people heads which are usually fairly visible, the majority of the other body parts are subject to heavy occlusion. This makes heads the only reliable discriminant feature in dense crowd counting and localization. Existing crowd counting and density estimation benchmark datasets such as UCF-CC-50 and ShanghaiTech  provide the heads centroid pixel location as labels. Conducting the crowd counting and density estimation as a regression task, seeks for regional isolation of the heads in the form of a binary mask. As the head size is subject to various factors such as camera specifications, point of view, perspective, distance and angle, generation of such mask could be challenging task, given the heads centroid pixel is the only provided form of annotation in existing benchmark datasets. 

The formation of the ground truth binary head masks in majority of the existing studies is either accomplished through a static two-dimensional Gaussian filter or a dynamic two-dimensional Gaussian filter paired with $k$ nearest neighbors approach. The static two-dimensional Gaussian filter assigns a fixed size Gaussian filter to each head regardless of the head size and proximity of the nearest neighbor. This approach does not attempt to compensate for crowd density, distance and camera perspective and incorporates significant amount of noise into ground truth data. The dynamic two-dimensional Gaussian filter approach employs the nearest neighbors search through $k$-d tree space partitioning approach, prioritizes the speed over integrity and does not deliver optimal results. In this approach the Gaussian filters are centered to the annotation points and spread based on the average euclidean distance among the three nearest neighbors. In both approaches, the spatial accumulation of all Gaussians creates the global density map for the given image. The following formula shows the commonly used dynamic two-dimensional Gaussian approach:
\begin{align}
D(x,f)=
\notag\\
\sum_{h=1}^{T}{\frac{1}{\sqrt{2\pi f(\sigma_h)}}}exp(-\frac{(x-x_h)^2 + (y-y_h)^2}{2f(\sigma_h)^2})
\end{align}
where $T$ is the total number of the heads presents in the given image, $\sigma_h$ is the sized for each head point positioned at $(x_h,y_h)$ determined by $k$-d tree space partitioning approach based on the the average euclidean distance among the three nearest neighbors and $f$ is a scaling constant. 

The dynamic Gaussian approach based on the $k$ nearest neighbors attempts to mitigate the crowd density, distance and camera perspective issues to some extent. However, this technique is not content aware and it introduces a significant amount of noise into the ground truth data, which in turn negatively affects the model's accuracy. Figure 1 shows some sample images from the ShanghaiTech dataset along with their respective ground truth density maps. It can be observed that both approaches are fairly unreliable and inconsistent in determining the true head sizes.

\begin{figure}
\captionsetup{font=small}
\centering
\includegraphics[width=7cm]{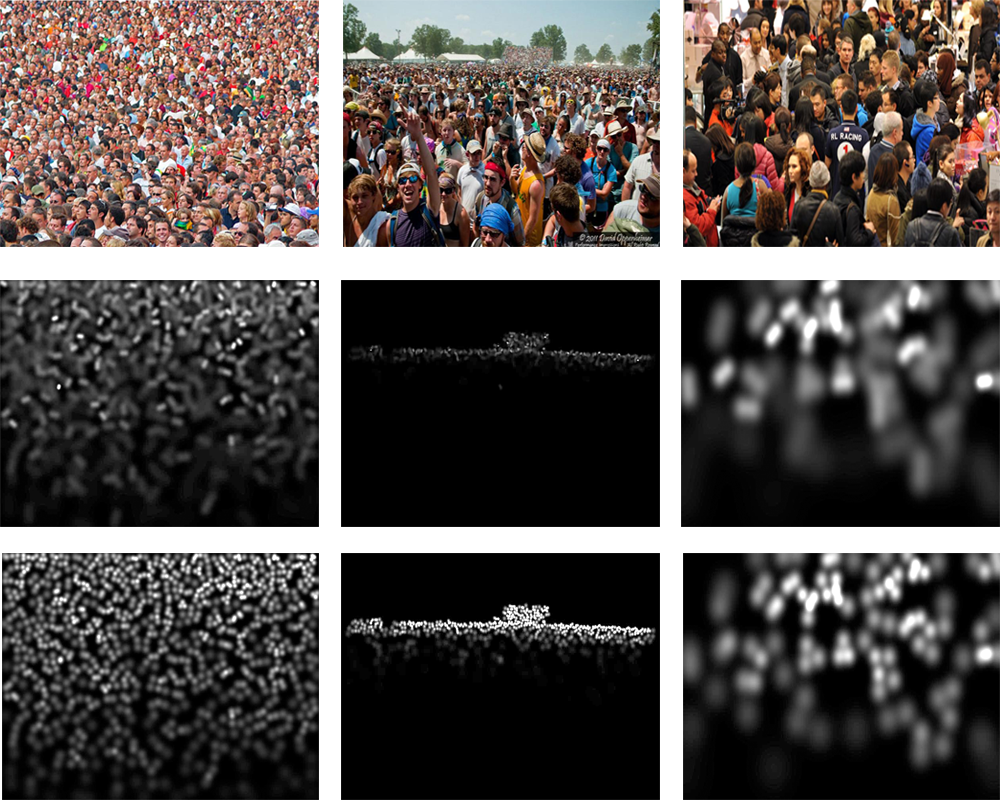}
\caption{
From top to bottom: sample images from the ShanghaiTech dataset, density map based on static two-dimensional Gaussian filter and density map based on dynamic two-dimensional Gaussian filter using  $k$-d tree space partitioning technique.}
\label{fig:ex}
\end{figure}

\section{Methodology}

In order to address the shortcomings of the existing ground truth density maps generation approaches, this study offers a content aware technique using combinations of Chan-Vese segmentation algorithm, two-dimensional Gaussian filter  and brute-force nearest neighbor search.

This technique is based on the Mumford-Shah functional for segmentation, and is widely used in the medical imaging field. The Chan-Vese segmentation algorithm is able to segment objects without prominently defined boundaries. This algorithm is based on level sets that are evolved iteratively to minimize an energy, which is defined by weighted values corresponding to the sum of differences intensity from the average value outside the segmented region, the sum of differences from the average value inside the segmented region, and a term which is dependent on the length of the boundary of the segmented region. As the head boundaries in highly dense crowds are not clearly defined, this technique can be used to segment the head regions from the background. Chan-Vese algorithm attempt to minimize the following energy function in an iterative process \cite{chan2001active}.
\begin{align}
F(c_1,c_2,G)=\mu . Len(G) + \nu . Area(in(G))
\notag\\
+\lambda_1 \int_{in(G)} |u_0(x,y)-c_1|^2 dxdy 
\notag\\
+\lambda_2 \int_{out(G)} |u_0(x,y)-c_2|^2 dxdy
\end{align}
where $G$ denote the initial head  which manually set to a $5x5$ bounding box centered on the labelled head point, $c_1$ will denote the average pixels' intensity inside the initial head region $G$, and $c_2$ denotes the average intensity of a square box, centered to the annotation head point and its boundary extended to the nearest neighbor head point. $\lambda_1$ , $\lambda_2$ and $\mu$ are positive scalars, manually set to 1, 1 and 0 respectively. A two-dimensional Gaussian filter which extends to the $G$ mean and centered to the head point is used to create the ground truth head mask.

Unlike $k$-d tree space partitioning technique which does not always delivers the absolute nearest neighbors, brute-force nearest neighbor search technique always guarantees to find the absolute nearest neighbors regardless of the distribution of the points. The brute-force nearest neighbor search technique does take considerably longer time ($O(n^2)$ vs $O(n\log{n})$) to find the nearest neighbors. However, since generating the ground truth density maps is a single-pass preliminary operation in crowd counting and density estimation, speed is a less of a priority. Since, the Chan-Vese segmentation algorithm only uses the very nearest neighbor head point to determine the boundary of the  outside region, the brute-force nearest neighbor search only looks for the very nearest head point. To create the global density map, we employed an exclusive cumulative of the Gaussians which addresses the head mask overlap issue. To maintain the count integrity, density map has been normalized at each iteration.

\section{Experimental Results}

In order to measure the effectiveness of our content-aware crowd density map generator, we have re-trained some of the notable state of the art deep models including Sindagi \textit{et al.} \cite{sindagi2017cnn} , Shi \textit{et al.} \cite{shi2018crowd} , Li \textit{et al.} \cite{li2018csrnet}  and Zhang \textit{et al.} \cite{zhang2015cross} using the density maps generated by the proposed  crowd density map generator. We have used the original implementation of these algorithms provided by authors in Github. All algorithms were trained and tested across both UCF-CC-50 and ShanghaiTech datasets using the proposed content-aware crowd density map generator as well as the commonly used existing ground truth density map generator. In some cases we were unable to reproduce the reported performance in the original manuscripts. However, as we were consistent with the experiments across both density map generators, validity and integrity of the comparison is not compromised. 

Table 1 shows the mean square error (MSE) comparison between the proposed and existing density map generator across ShanghaiTech dataset part A and B. It can be observed that using the proposed content-aware density map generator, MSE has been consistently decreased across relatively all investigated models. The improvements is more pronounced in ShanghaiTech part A dataset. ShanghaiTech part A dataset exhibits more challenging and dynamic crowd scenarios. The results convey the proposed method could deliver better depiction of the ground truth density maps. Table 2 compares the MSE and mean absolute error (MAE) between the proposed and existing density map generator using extremely challenging UCF-CC-50 dataset. Similar to the results in ShanghaiTech dataset, there is a notable improvement in both MSE and MAE metrics. 
 
Figure 2 compares the density maps generated using the existing approach based on $k$-d tree space partitioning technique and the proposed content-aware crowd density map generator. It can be observed that in highly dense crowds, the proposed method generates more  more granular density maps with lesser overlaps between neighbor Guassians. The proposed method uses combination of pixels intensity and nearest neighbors to adjust the size of the Guassians per head. Figure 2 shows this technique significantly improves the integrity of the density map relative to the input image.

\begin{table}[]
\captionsetup{font=small}
\caption{MSE comparison between the proposed and existing density map generator across ShanghaiTech dataset}
\label{tab:my-table}
\resizebox{7.35cm}{!}{%
\begin{tabular}{@{}lllll@{}}
\toprule
 & \multicolumn{2}{c}{\begin{tabular}[c]{@{}c@{}}Existing Density map \\ Generator\end{tabular}} & \multicolumn{2}{c}{\begin{tabular}[c]{@{}c@{}}Proposed Density map\\  Generator\end{tabular}} \\ \midrule
Method & \multicolumn{1}{c}{\begin{tabular}[c]{@{}c@{}}ShTech-A\end{tabular}} & \multicolumn{1}{c}{\begin{tabular}[c]{@{}c@{}}ShTech-B\end{tabular}} & \multicolumn{1}{c}{\begin{tabular}[c]{@{}c@{}}ShTechA\end{tabular}} & \multicolumn{1}{c}{\begin{tabular}[c]{@{}c@{}}ShTechB\end{tabular}} \\
Sindagi \textit{et al.} & 152 & 31 & 149 & 28 \\
Shi \textit{et al.} & 112 & 26 & 110 & 26 \\
Li \textit{et al.} & 115 & 16 & 113 & 16 \\
Zhang \textit{et al.} & 197 & 66 & 191 & 57 \\ \bottomrule
\end{tabular}%
}
\end{table}

\begin{table}[]
\captionsetup{font=small}
\caption{MSE and MAE comparison between the proposed and existing density map generator across UCF-CC-50 dataset}
\label{tab:my-table}
\resizebox{7.35cm}{!}{%
\begin{tabular}{@{}lllll@{}}
\toprule
 & \multicolumn{2}{c}{\begin{tabular}[c]{@{}c@{}}Existing Density map \\ Generator\end{tabular}} & \multicolumn{2}{c}{\begin{tabular}[c]{@{}c@{}}Proposed Density map\\  Generator\end{tabular}} \\ \midrule
Method & \multicolumn{1}{c}{\begin{tabular}[c]{@{}c@{}}UCF-CC-50\\ MSE\end{tabular}} & \multicolumn{1}{c}{\begin{tabular}[c]{@{}c@{}}UCF-CC-50\\ MAE\end{tabular}} & \multicolumn{1}{c}{\begin{tabular}[c]{@{}c@{}}UCF-CC-50\\ MSE\end{tabular}} & \multicolumn{1}{c}{\begin{tabular}[c]{@{}c@{}}UCF-CC-50\\ MAE\end{tabular}} \\
Sindagi \textit{et al.} & 397  & 322 & 397 & 320 \\
Shi \textit{et al.} & 415 & 293 & 414 & 286 \\
Li \textit{et al.} & 397 & 266 & 396 & 264 \\
Zhang \textit{et al.} & 498 & 467 & 483 & 459 \\ \bottomrule
\end{tabular}%
}
\end{table}

\begin{figure}
\captionsetup{font=small}
\centering
\includegraphics[width=7.3cm]{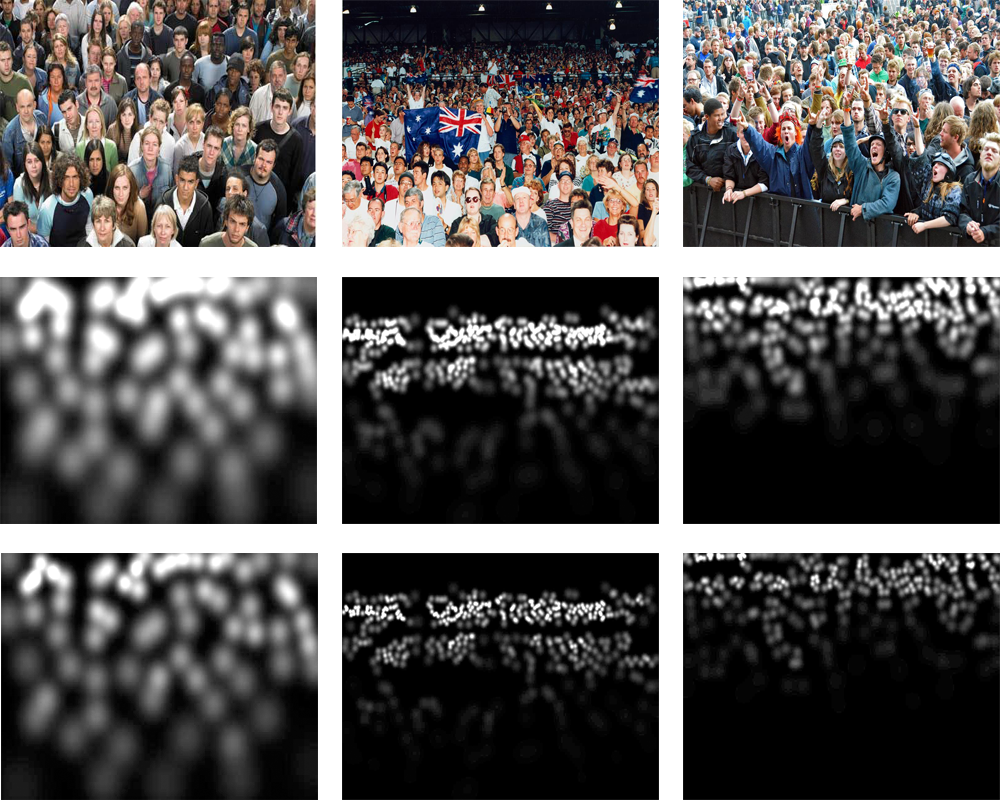}
\caption{
From top to bottom: sample images from ShanghaiTech dataset, density map generated using the existing method and density map generated using the proposed method}
\label{fig:ex}
\end{figure}

\section{Conclusion}

Creating an accurate model for crowd counting and density estimation demands for a large and highly reliable ground truth data in the first place. However, the existing crowd counting and density estimation benchmark datasets are not only limited in terms of size, but also lack in terms of annotation methodology. This study attempted to address this issue through a content-aware technique which employed combinations of Chan-Vese segmentation algorithm, two-dimensional Gaussian filter and brute-force nearest neighbor search to generate the ground truth density maps. Experiment results shows by replacing the commonly practiced ground truth density map generators with the proposed content-aware method, the existing state of the art crowd counting models can achieve higher level of count and localization accuracy.

\section*{Acknowledgments}
This work is co-funded by the EU-H2020 within the MONICA project under grant agreement number 732350. The Titan X Pascal used for this research was donated by NVIDIA.

\bibliographystyle{unsrt}
\bibliography{example.bib}

\end{document}